\documentclass{article}

\usepackage{microtype}
\usepackage{graphicx}
\usepackage{subfigure}
\usepackage{placeins}
\usepackage{booktabs} 

\usepackage{hyperref}

\usepackage[accepted]{icml2020}

\icmltitlerunning{Multi-Image Steganography Using Deep Neural Networks}

\begin{document}

\twocolumn[
\icmltitle{Multi-Image Steganography Using Deep Neural Networks}



\icmlsetsymbol{equal}{*}

\begin{icmlauthorlist}
\icmlauthor{Abhishek Das}{CMU}
\icmlauthor{Japsimar Singh Wahi}{CMU}
\icmlauthor{Mansi Anand}{CMU2}
\icmlauthor{Yugant Rana}{CMU2}
\end{icmlauthorlist}

\icmlaffiliation{CMU}{Electrical and Computer Engineering, Carnegie Mellon University}
\icmlaffiliation{CMU2}{Information Networking Institute, Carnegie Mellon University}



\vskip 0.3in
]



\printAffiliationsAndNotice{}  


\section{Abstract}
\label{submission}
Steganography is the science of hiding a secret message within an ordinary public message. Over the years, steganography has been used to encode a lower resolution image into a higher resolution image by simple methods like LSB manipulation. We aim to utilize deep neural networks for the encoding and decoding of multiple secret images inside a single cover image of the same resolution.

\section{Introduction}
\label{submission}

Steganography refers to the technique to hide secret messages within a non-secret message in order to avoid detection during message transmission. The secret data is then extracted from the encoded non-secret message at its destination. The use of steganography can be combined with encryption as an extra step for hiding or protecting data. Traditionally, steganography is performed to embed low-resolution images onto a high-resolution image using naive methods like LSB manipulation.

\textbf{Motivation for the project} comes from the recent works, like that of \cite{NIPS2017_6802}, \cite{Hayes2017GeneratingSI}, and \cite{DBLP:journals/corr/abs-1807-09937}. These papers suggest the use of deep neural networks to model the data-hiding pipeline. These methods have significantly improved the efficiency in terms of maintaining the secrecy and quality of the encoded messages. Recently, similar work in terms of audio signal steganography, like \cite{kreuk2019hide}, has shown that deep neural networks can be used to encode multiple audio messages onto a single cover message. 

\textbf{We aim to make an effort in a similar direction, by utilizing the ideas from the aforementioned papers to encode multiple images into the single cover image.} Unlike traditional methods, we use the same resolution cover and secret images and we aim to keep the changes to the encoded cover image unnoticeable to human perception and statistical analysis, while at the same time keeping the decoded images highly intelligible.

The scripts have been made publicly available to the research community for further development
 \href{https://github.com/JapsimarSinghWahi/DeepSteganography}{here}.

In what follows, we discuss the related prior work for such a problem in the next section (3), followed by Baseline Implementations in section (4) Datasets in section (5), our Proposed Methodology in section (6) followed with its Results and Discussion in section (7). Finally, we end the discussion with Future Directions, Conclusion and Acknowledgement in the sections (8), (9), and (10) respectively.

\section{Related Work}

Out of the several implementations, below two are most aligned and important to our goal.

\subsection{Hiding Images in Plain Sight: Deep Steganography}

\begin{figure*}[t]
\vskip 0.2in
\begin{center}
\centerline{\includegraphics[width=\textwidth]{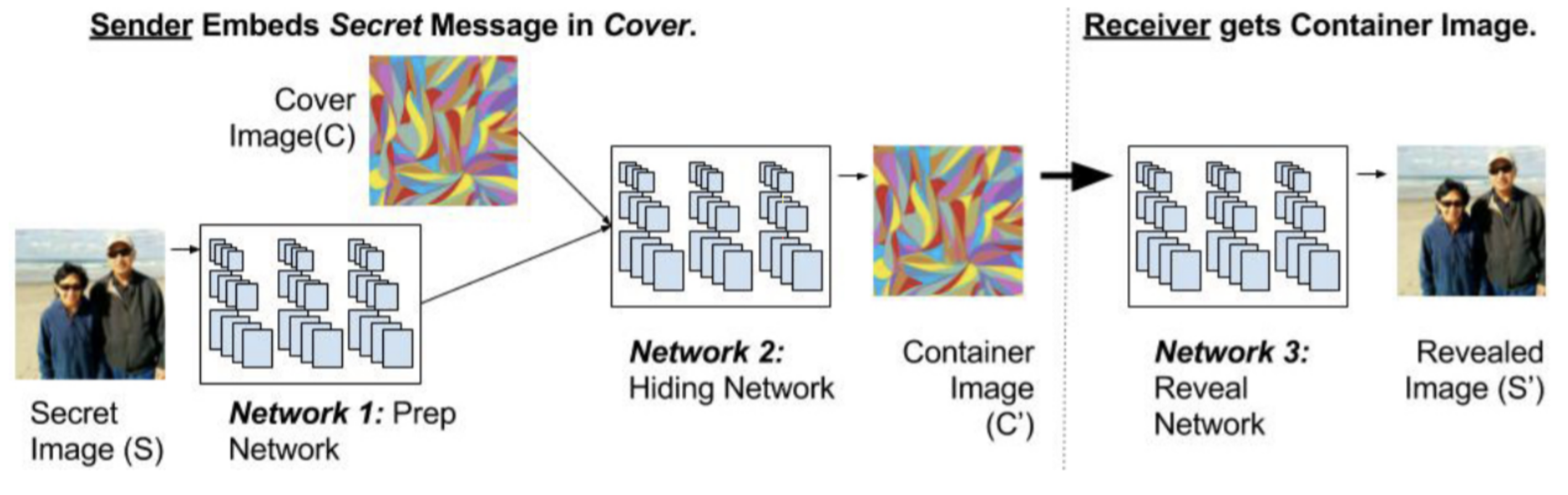}}
\caption{The three components of the full system. Left: Secret-Image preparation. Center: Hiding the image in the cover image. Right: Uncovering the hidden image with the reveal network; this is trained simultaneously, but is used by the receiver.}
\label{fig:baluja_arch}
\end{center}
\vskip -0.4in
\end{figure*}

 \cite{NIPS2017_6802} attempts to place a full-sized color image within another image of the same size. Deep neural networks are simultaneously trained to create the hiding and revealing processes and are designed to specifically work as a pair. The system is trained on images drawn randomly from the ImageNet database and works well on natural images from a wide variety of sources.  Unlike many popular steganographic methods that encode the secret message within the least significant bits of the cover image, their approach compresses and distributes the secret image’s representation across all of the available bits.

The three components involved in the system include-
    \begin{enumerate}{}
    \item Preparation Network - prepares the secret image to be hidden. If the secret-image (size M$\times$M) is smaller than the cover image (N$\times$N), the preparation network progressively increases the size of the secret image to the size of the cover, thereby distributing the secret image’s bits across the entire N $\times$ N pixels.
    \item Hiding Network - takes as input the output of the preparation-network and the cover image, and creates the Container image. The input to this network is an N $\times$ N pixel field, with depth concatenated RGB channels of the cover image and the transformed channels of the secret image.
    \item Reveal Network - used by the receiver of the image; it is the decoder. It receives only the Container image (neither the cover nor the secret image). The decoder network removes the cover image to reveal the secret image.
    \end{enumerate}
    
The paper by \cite{NIPS2017_6802} talks about how a trained system must learn to compress the information from the secret image into the least noticeable portions of the cover image. However, no explicit attempt has been made to actively hide the existence of that information from machine detection. They trained the steganalysis networks as binary classifiers, using the unperturbed ImageNet images as negative samples, and their containers as positive examples. The paper serves a baseline for single secret image encoding. However, it does not talk about multi-image steganography.

\begin{figure}[h]
\vskip 0.2in
\begin{center}
\centerline{\includegraphics[width=\columnwidth]{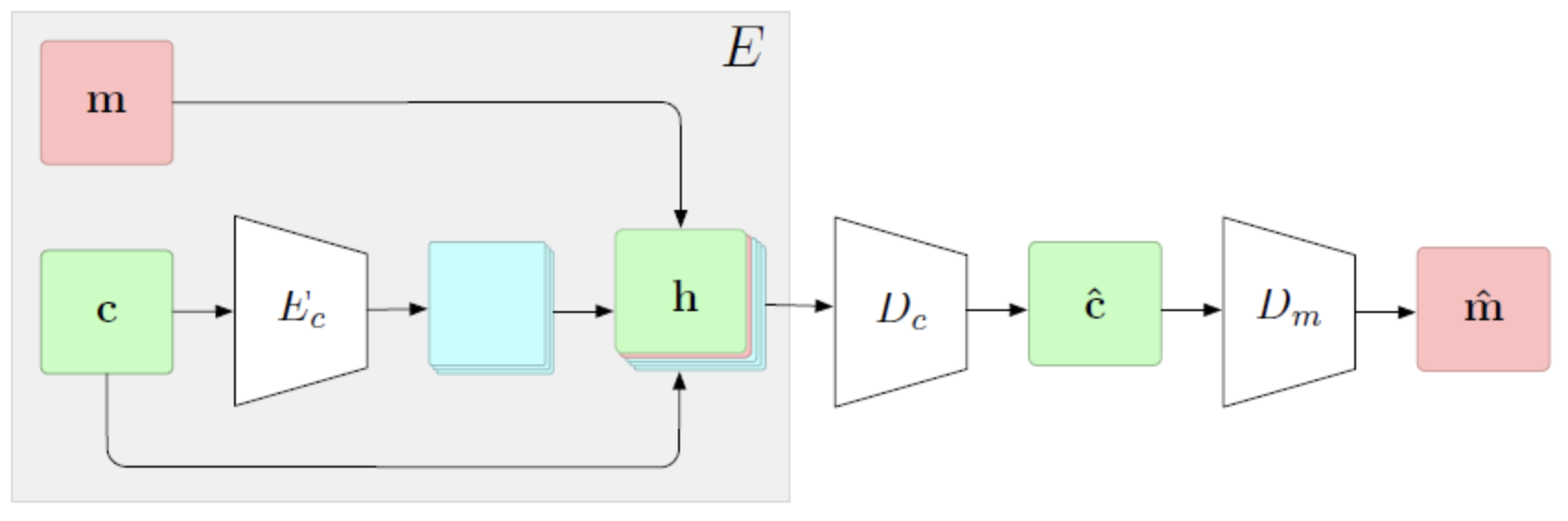}}
\caption{Model overview: the encoder E gets as input the carrier c and the message m, it encodes c using the carrier encoder $E_c$ and concatenates $E_c$(c) to c and m to generate h. Then, the decoder carrier $D_c$ generates the new encoded carrier, from which the decoder message $D_m$ decodes the message $\hat{m}$ . During training the reconstruction loss is applied between c and m to $\hat{c}$ and $\hat{m}$, respectively}
\label{fig:kreuk_arch}
\end{center}
\vskip -0.4in
\end{figure}

\subsection{Hide and Speak: Deep Neural Networks for Speech Steganography}

\cite{kreuk2019hide} implements steganography for speech data using deep neural networks. It is based on an architecture that comprises 3 subparts, i.e, Encoder Network, Carrier Decoder Network, and a Message Decoder Network. They utilize ideas from \cite{DBLP:journals/corr/abs-1807-09937} to extend the encoder network to audio signals. The architecture of the model comprises of 3 sub-parts:
    \begin{enumerate}{}
        \itemsep0em 
        \item An Encoder Network ($E_c$)
        \item A Carrier Decoder Network ($D_c$)
        \item A Message Decoder Network ($D_m$)
    \end{enumerate}

In the Carrier/cover encoder network, the encoded carrier ($E_c$(c)) is appended with the carrier (c) and the secret message (m), forming, [$E_c$(c);c;m]. This output is fed to the Carrier Decoder ($D_c$) which outputs the carrier embedded with a hidden message. Finally, this is fed to the Message Decoder ($D_m$) which reconstructs the hidden message.

The first part learns to extract a map of potential redundancies from the carrier signal. The second part utilizes the map to best “stuff” a secret message into the carrier such that the carrier is minimally affected. The part third learns to extract the hidden message from the steganographically-modified carrier. All the components in these networks are Gated Convs. $E_c$ is composed of 3 blocks of Gated Convs, $D_c$ 4 and $D_m$ 6 blocks of Gated Convs. Each block contains 64 kernels of size 3*3.

This paper demonstrates the capability to hide multiple secret messages in a single carrier, which aligns with our goals. In the paper, five independent speech messages have been hidden in a single speech recording. This is achieved by 2 different approaches. One approach utilizes multiple decoders, with each decoder trained to decode a different message. The other approach utilizes a conditional decoder that also takes in as input a code indicating the message index to be encoded. 

\begin{figure*}[t]
\vskip 0.2in
\begin{center}
\centerline{\includegraphics[width=\textwidth]{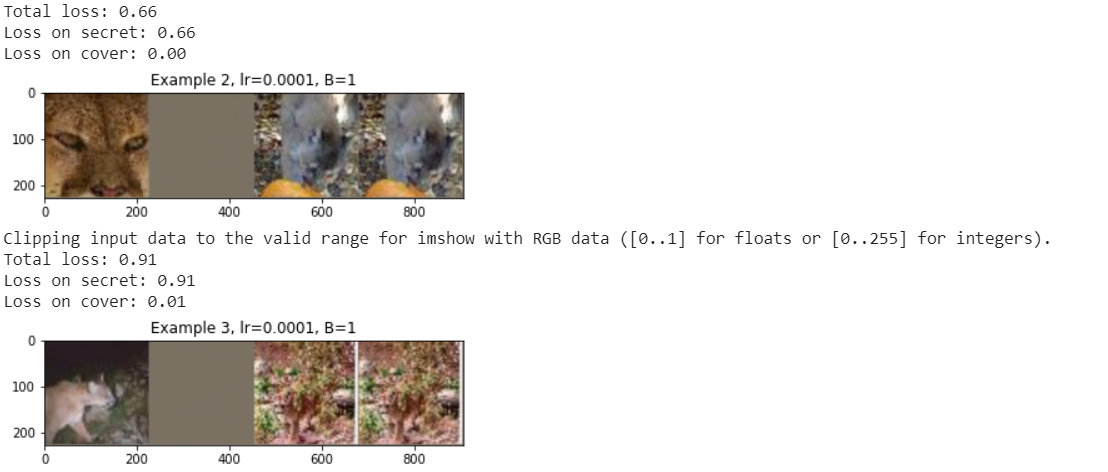}}
\caption{The following results shows the cover and hidden images before and after running the model for the first 2 images after 20 epochs.}
\label{fig:fpingham_res1}
\end{center}
\vskip -0.4in
\end{figure*}

We borrowed the concept of multiple decoders from this paper and used it to fetch multiple secret images from the cover image which looks like the cover image but consists of the secret images being hidden inside the cover image via passing through separate prep networks and then concatenated together. For this, we utilize the loss as defined by this paper for the multiple decoders. For our case, we extend the loss function defined in this paper for our use case. We take the reveal loss for each decoder as
\begin{center}
$Reveal Loss = \lambda_s*\sum||S- S'||^2$
\end{center}
and for the entire system as summation of reveal losses for each decoder added with the loss calculated for cover image as well.

$L(c,M) = \lambda_c*||C - D_c(E(c,M))||^2 + \lambda_m*\sum||m_i - D_m(D_c(E(c,M)))||^2$
, where $ M= {m_i}^k$

We used this approach to introduce multi-image steganography using the above idea from this paper and extending it to images to add multiple images in one cover image and then retrieving it.

\begin{figure*}
\vskip 0.2in
\begin{center}
\centerline{\includegraphics[width=\textwidth]{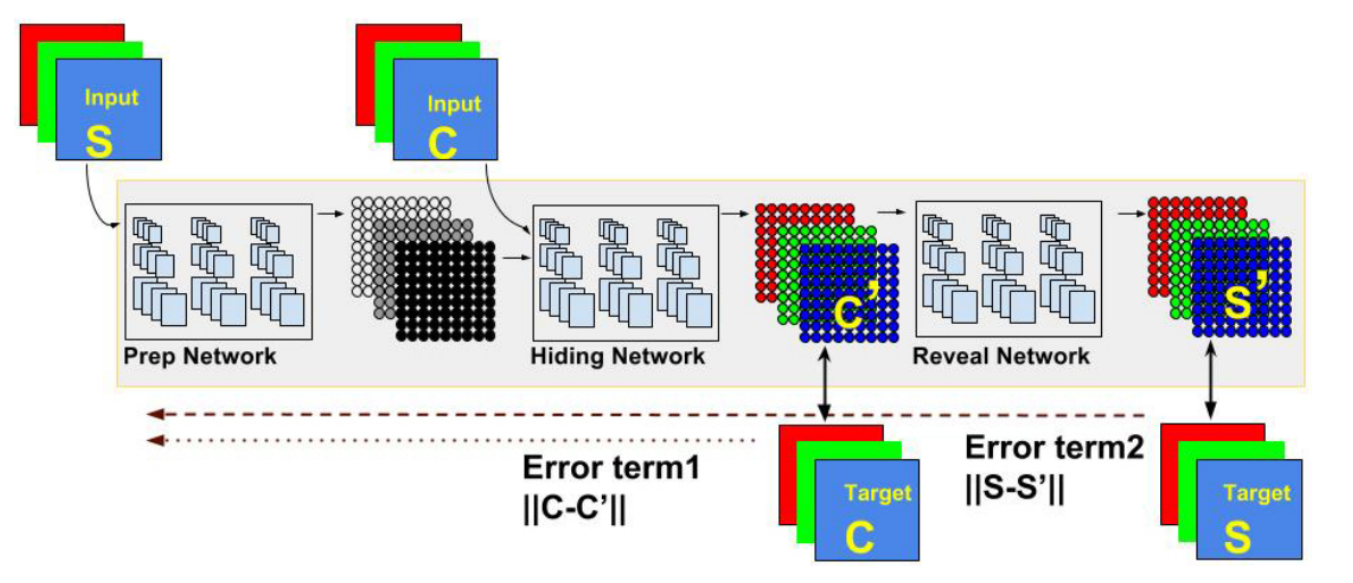}}
\caption{Baluja's implementation by Alexandre}
\label{fig:Alexandre_arch}
\end{center}
\vskip -0.4in
\end{figure*}

\section{Baseline Implementations}

We aimed to implement a baseline single image steganography model over which we could preform out extensions. Since \cite{NIPS2017_6802} has several implementations, we implemented two of the most popular implementations and analyzed them to be suitable for the extensions for our model. The details of these implementations are as follows:

\subsection{{Ingham's Implementation}}

\cite{fpingham} is a PyTorch based implementation which follows the architecture shown in Figure \ref{fig:baluja_arch}. The architecture includes a Prep Network, a Hidden Network, and a Reveal Network and embeds a single secret image onto a single cover image. The model architecture is defined as follows: 
\begin{enumerate}
    \item \textbf{Prep Network} - Two sets of three sequential layers consisting of (Conv2d - Relu) combinations, concatenated and fed into the next set.
    \item \textbf{Hidden Network} - Similar to the prep network above but includes an extra Conv2D sequential layer for adding Gaussian noise to the cover. This allows the hidden information to be encoded in bits other than the LSB of the cover image.
    \item \textbf{Reveal Network} - Similar to the above networks, with an extra Conv2d at the end.
 \end{enumerate}

\subsubsection{Implementation Details}
For full code please refer \href{https://drive.google.com/open?id=1lzJdYhmjusQxJKZQTnoGU9WUqsaEvxIC}{\color{blue}{here}}. The model is explained below:
\begin{enumerate}
\itemsep0em 
\item Transformations: Scaling, random crop and normalization
\item Optimizer: Adam, with learning rate 0.001
\item Customized Loss, as suggested by \cite{NIPS2017_6802}. coverloss + $\beta \times$hiddenloss)
\end{enumerate}

\subsubsection{Results}

The author of the implementation had shown desirable results for higher resolution images. Since we used low-resolution images with an equal resolution for both secret and cover, we did not see similar results in our implementations. See Figure \ref{fig:fpingham_res1}. The implementation generated lossy secret images, [left-side images in results] while retaining the cover image almost completely [right-side images in results]. Since other baseline models performed significantly better, we decided not to proceed with this implementation.

\subsection{{ Alexandre's Implementation}} 
Alexandre's is another implementation \cite{alexandre} of Baluja's paper, based on Keras. There are new features introduced in this model that others have not implemented like noise addition, mean stabilization. We have tried various models to see which configuration provides the best results and is fastest. The model architecture has three parts:
\begin{enumerate}
    \item \textbf{Preparation Network} : Transforms secret image to be concatenated with cover.
    \item \textbf{Hiding Network}: Converts the concatenated image into encoded cover.
    \item \textbf{Reveal Network}: Extracts the secret image from the encoded cover.
\end{enumerate}

Hiding and reveal networks use aggregated Conv2D layers: 5 layers of 65 filters [50 3x3 filters, 10 4x4 filters, and 5 5x5 filters]. Prep network uses 2 layers of similar structure. All Conv 2D layers are followed by ReLU activation.


\subsubsection{Implementation Details}
For full code please refer \href{https://colab.research.google.com/drive/1_-k_0NpGouhHurBTZyF-jiTruUHJ2K3Y?usp=sharing}{\color{blue}{here}}.
Model is explained below:
\begin{enumerate}
    \item Adam optimizer, with learning rate as 0.001 and a custom scheduler.
    \item Model has been trained for 800 epochs with a batch size of 256 and an additional 100 epochs with batch size of 32. 
    \item To make sure weights are updated only once, reveal network weights are frozen before adding it to the full model.
    \item Gaussian noise with 0.01 standard deviation is added to the encoder's output before passing it through the decoder. 
    \item Mean sum of squared error has been used for calculating the decoder's loss.
\end{enumerate}

\subsubsection{Results}
\begin{figure*}[t]
\vskip 0.2in
\begin{center}
\centerline{\includegraphics[width=\textwidth]{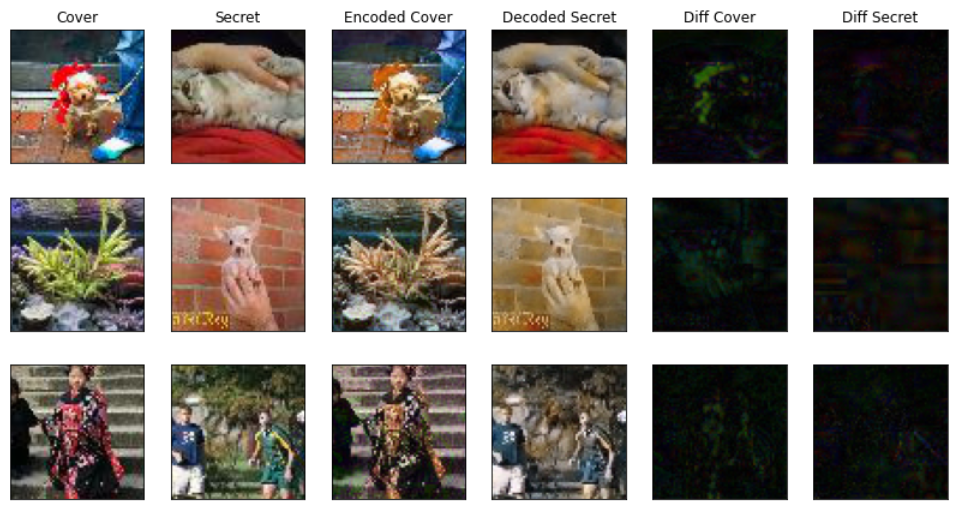}}
\caption{The results showing the cover and hidden images before and after running the model for 900 epochs. Left to Right Columns are: Cover Image, Secret Image, Encoded Cover Image, Decoded Secret Image, Diff Cover Image, Diff Secret Image. We can notice than the differences between the original cover and the encoded cover is almost going null. Same with the original secret and the decoded secret image.}
\label{fig:Alexandre-Result_5}
\end{center}
\vskip -0.4in
\end{figure*}

As it can be seen in Figure \ref{fig:Alexandre-Result_5}, the model generates decent results in the Tiny ImageNet dataset. The generated images showed a minimal loss for both the cover and the secret images. Owing to the performance of the model with a single image steganography we decided to use this implementation for our work going ahead.

\section{Datasets}

    Since our model does not have specific requirements pertaining to the classes of the images, we used the \textbf{Tiny ImageNet} \cite{tiny_imagenet} dataset in order to obtain the secret and cover images. 
    The dataset is the collection of 64$\times$64$\times$3 images, used by the Stanford CS231 class. Further extensions of the final model can also be applied to larger images from datasets like ImageNet \cite{imagenet_cvpr09}. We have also used Tiny ImageNet for faster training. 
    
    Our training set is made of a random subset of images from all 200 classes. 2000 images are randomly sampled. The image vectors are normalized across RGB values. We split the entire training data into four halves, one for the cover image and the other three halves for three secret images.

\begin{figure*}[t]
\vskip 0.2in
\begin{center}
\centerline{\includegraphics[width=\textwidth]{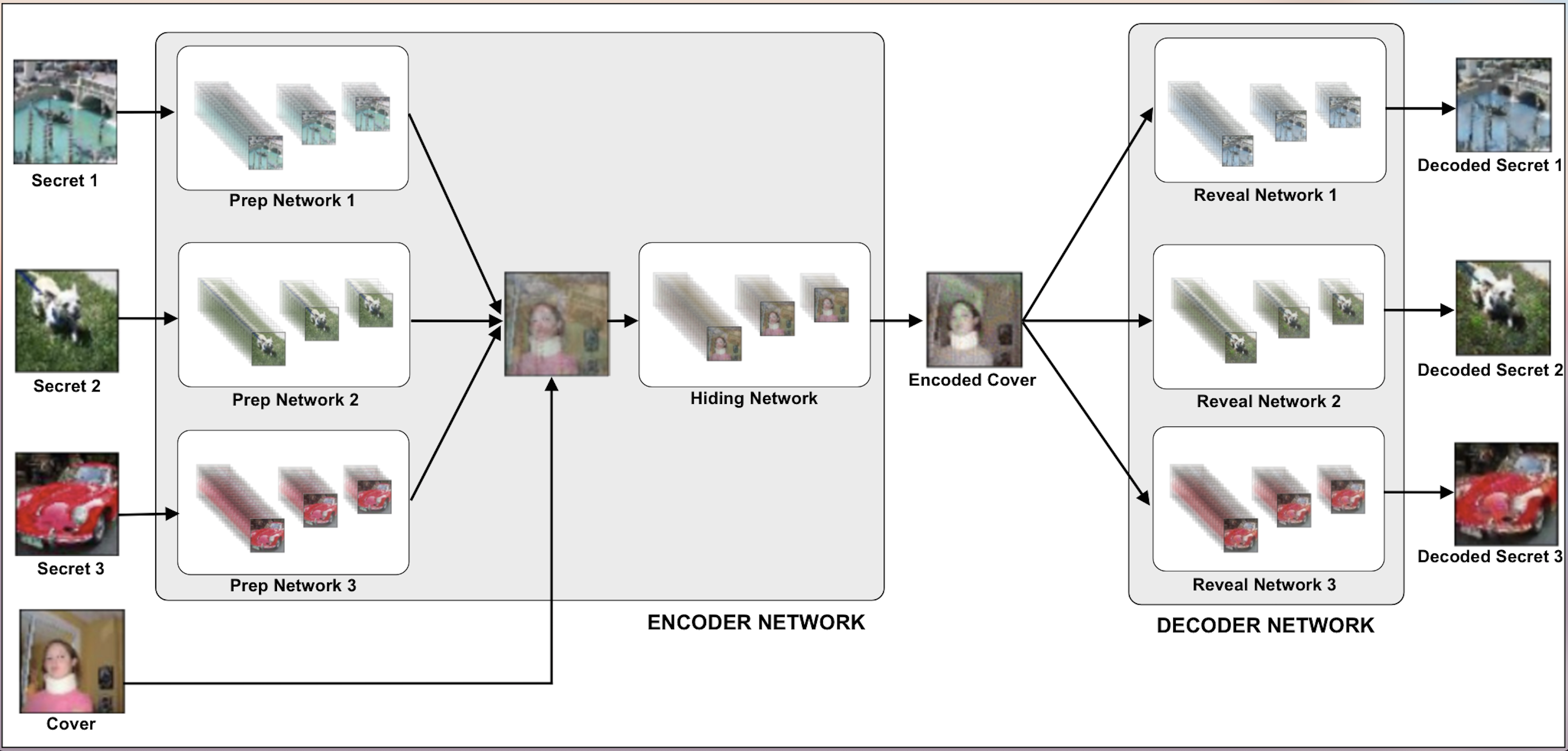}}
\caption{The image shows the DeepSteg architecture with multiple CNN based sub-networks. Inside the encoder, the prep networks convert the input secret images into images that can be concatenated to the cover. The concatenation is then passed through the hiding network to generate the encode cover. In the decoder network, separate reveal networks are deployed to generate the decoded secrets out of the encoded cover.}
\label{fig:StegNet}
\end{center}
\vskip -0.2in
\end{figure*}

\section{Proposed Methodology}
We aim to perform multi-image steganography, hiding three or more images in a single cover image. The embedded secret images must be retrievable with minimum loss. The encoded cover image must look like the original cover image. To perform this, we combine the idea of \cite{NIPS2017_6802} and \cite{kreuk2019hide}. We take the network implementation idea of having a prep and hiding network as an encoder and a reveal network as a decoder from \cite{NIPS2017_6802}. To extend this for multiple images, we pass multiple secret images via the prep network and then concatenating these resulting data with the carrier image and then finally send this via the Hiding network. We then take the idea of having multiple decoders, one per secret image, from \cite{kreuk2019hide} to retrieve all the secret images from the container image.To improve the security of our image retrieval model, we extend the idea presented by \cite{NIPS2017_6802} of putting secret images with noise in the original cover image instead of putting the secret images at the LSBs of the original cover image.

\textbf{Using of Multiple Prep and Reveal Networks}. In their implementation for multiple audio signal steganography, \cite{kreuk2019hide}, suggested the use of multiple decoders to derive the decoded secret from a single encoded cover. This technique is an extension of the same idea in the image domain. It does not require scaling down the size of the image or sacrificing on the color channels of the secret images. 

There was another technique of using conditional decoders instead of multiple decoders but in this implementation, we have only worked on implementing multiple prep/reveal networks. Hence, we decided to build our extension based on this technique. A brief description of the encoder/decoder framework as per this technique is as follows:

\begin{itemize}
\item \textbf{ENCODER}: Consists of multiple prep networks, each corresponding to separate secret image input. Prep network outputs are concatenated together with the cover image and then fed through the Hiding network.
\item \textbf{DECODER}: The decoder network comprises of multiple reveal networks, each of which is trained separately to decode its corresponding message.
\end{itemize}

\subsection{Model Architecture}
The model architecture is as described in the Figure \ref{fig:StegNet}. The underlying architecture of each of the sub-networks is as follows:

\begin{enumerate}
    \item \textbf{Prep Networks}: Each of the prep networks consists of the aggregation of 2 layers. With each of the layers made up of 3 separate Conv2D layers. The number of channels for these 3 Conv2D layers is 50, 10, and 5 respectively with the kernel sizes like 3, 4, and 5 for each layer. The stride length constantly remains 1 along both the axis. Appropriate padding is added to each Conv2D layer so as to keep output image in the same dimensions. Each Conv2d layer is followed with a ReLU activation.
    
    \item \textbf{Hiding Network}: The hiding network is an aggregation of 5 layers. With each of these layers made up of the 3 separate Conv2D layers. The underlying structure of the Conv2D layers in the hiding network is similar to the Conv2D layers in the Prep Network.
    
    \item \textbf{Reveal Network}: Each of the reveal networks shares a similar underlying architecture with the hiding network, using 5 layers of similarly formed Conv2D layers.
\end{enumerate}

\begin{figure*}[t]
\vskip 0.2in
\begin{center}
\centerline{\includegraphics[width=\textwidth]{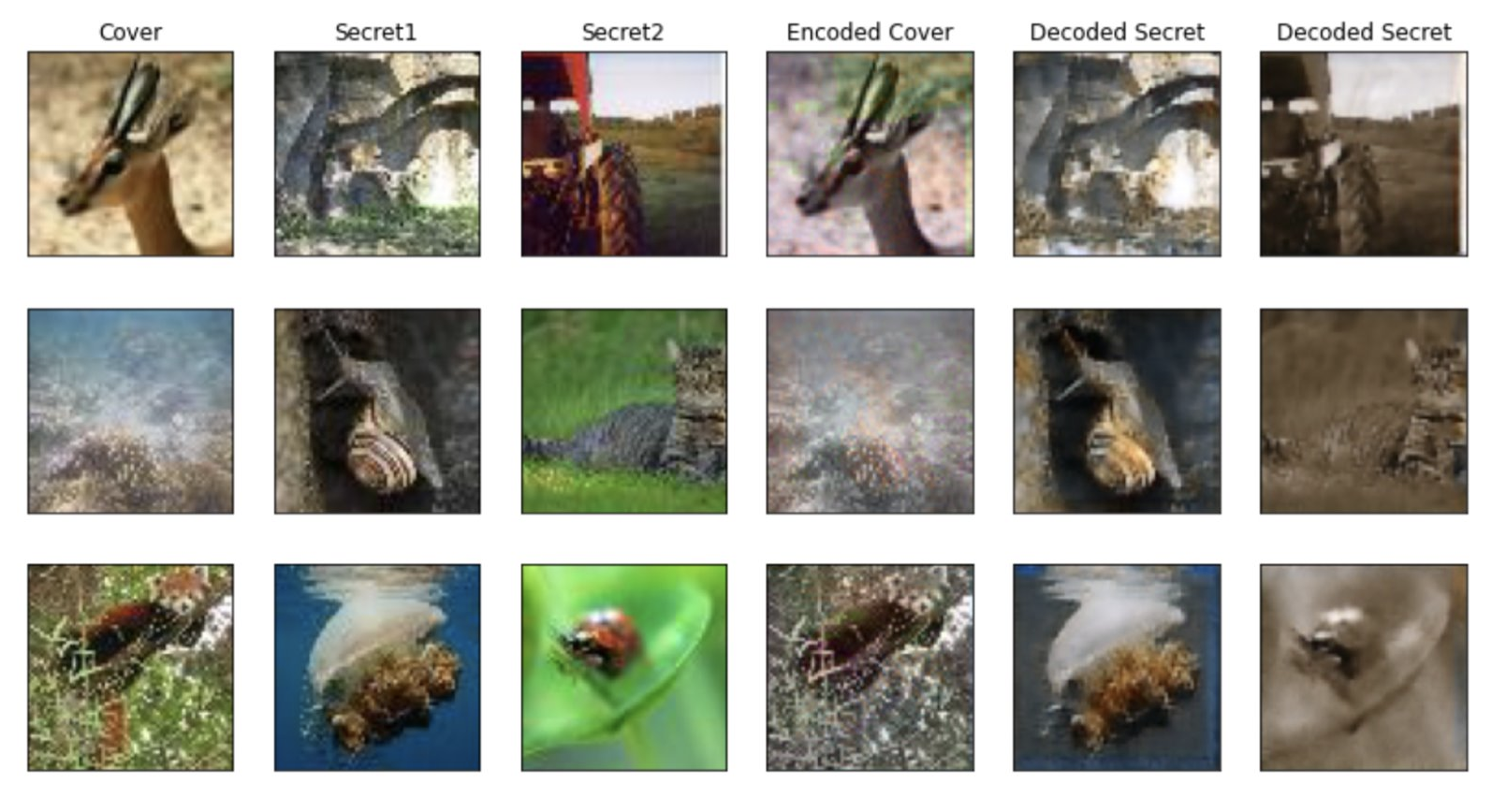}}
\caption{Result of hiding two secret images. Left to Right Columns are: Cover Image, Secret Image1, Secret Image2, Encoded Cover Image, Decoded Secret Image1, Decoded Secret Image2.}
\label{fig:2_Image_result}
\end{center}
\vskip -0.2in
\end{figure*}

\subsection{Implementation Details}
For full code please refer \href{https://colab.research.google.com/drive/1CuvpGxHtLUu-2HnlQu5hApSY4RRbbNJ4?usp=sharing}{\color{blue}{here}}.
The training details are explained below:
\begin{enumerate}
    \item Adam optimizer has been used with a custom LR scheduler.
    \item Learning rate remains constant with 0.001 till first 200 epochs, decreasing to 0.0003 from 200 epochs to 400 epochs and further decreasing it 0.00003 for remaining iterations.
    \item Model has been trained for 750 epochs with a batch size of 256 and an additional 400 epochs with a batch size of 32. 
    \item Tiny Image Dataset has been used, where images are 64x64. Dataset is created by taking 10 images per class for train and 2000 images in total for train and test.
    \item Train set is divided into 2 sections. First 1000 images are used for training as secret images and rest 1000 for cover images.
    \item Preparation and Hiding networks share the same stacked Keras model and loss. Reveal network has its own stacked model and its own loss function. 
    \item Currently, the learning rate is 0.001. 
    \item To make sure weights are updated only once, reveal network weights are frozen before adding it to the full model.
    \item Gaussian noise with 0.01 standard deviation is added to encoder's output before passing it through the decoder.
    \item Mean sum of squared error has been used for calculating decoder's loss.
    \item The loss used is for the full model is represented as:
    \begin{center}
        $Loss = \lambda_c*||C-C'||^2 + \lambda_s*||S_1 - S_1'||^2 + \lambda_s*||S_2 - S_2'||^2 + \lambda_s*||S_3 - S_3'||^2$
    \end{center}
    \item While training the reveal network we only consider the secret image component of the loss.
    \item During the full model training, the loss for both cover and secret image is taken into consideration.
    \item Currently we are taking both the $\lambda_s$ and $\lambda_c$ as 1.0.
\end{enumerate}


\begin{figure*}[t]
\vskip 0.2in
\begin{center}
\centerline{\includegraphics[width=\textwidth]{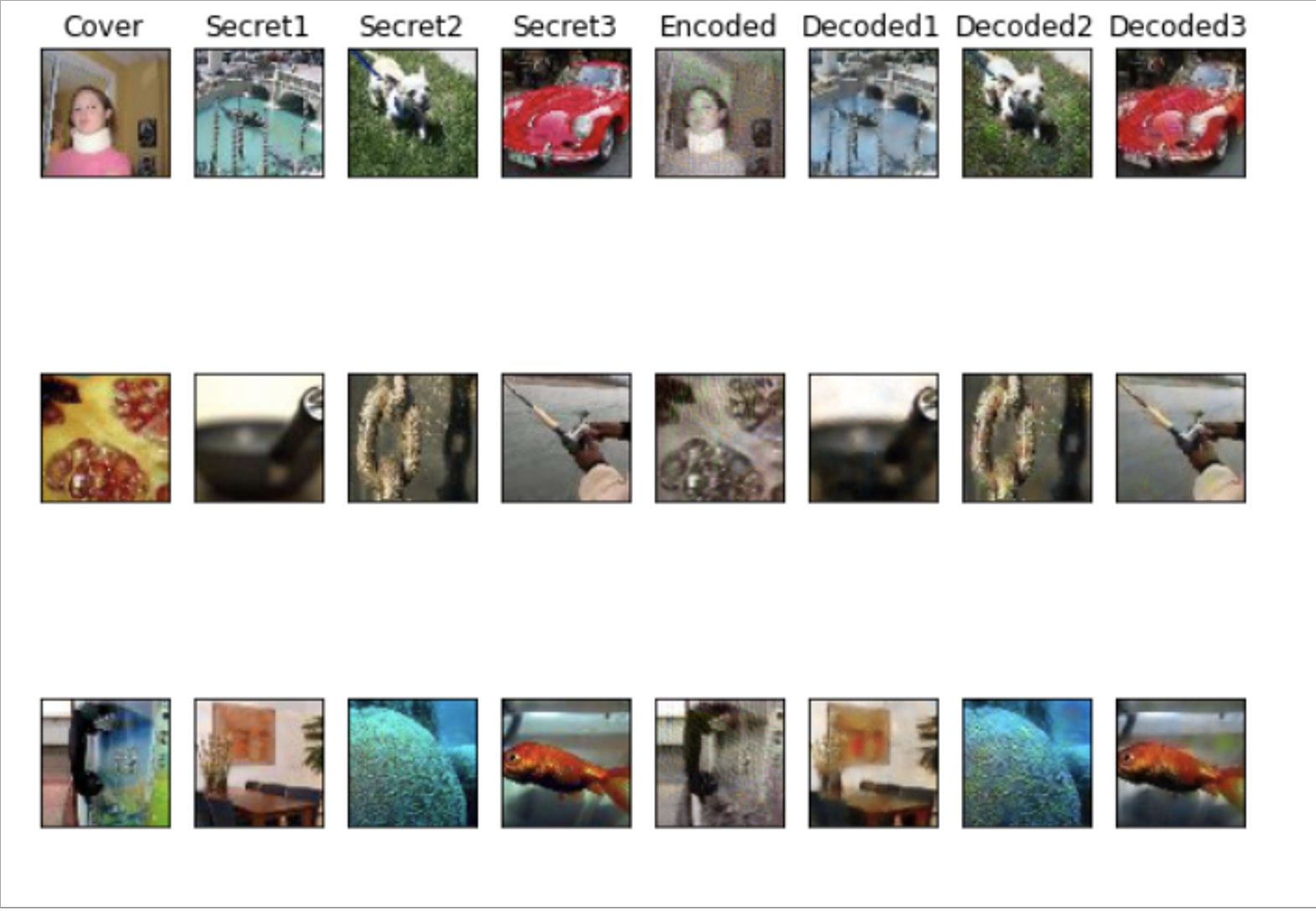}}
\caption{Result of hiding three secret images. Left to Right Columns are: Cover Image, Secret Image1, Secret Image2, Secret Image2, Secret Image3,Encoded Cover Image, Decoded Secret Image1, Decoded Secret Image2, Decoded Secret Image3.}
\label{fig:3_Image_result}
\end{center}
\vskip -0.2in
\end{figure*}

\section{Results and Discussion}
Figure \ref{fig:2_Image_result} depicts the results of hiding two secret images over a single cover image. The input images are depicted on the left side while the encoder/decoder outputs are presented on the left hand size. The encoded cover image looks similar to the original cover to a great extent, and it does not reveal information about the secret images. 

The results of hiding three secret images are shown in Figure \ref{fig:3_Image_result}. The encoded cover is more lossy as compared to the case when only two secret images are used. The secret images are retrieved successfully in both cases.

The losses received for the below results after 750 epochs were as below - 
\begin{enumerate}
\itemsep0em 
\item Loss of Entire Setup - 182053.70
\item Loss secret1 - 51495.24
\item Loss secret2 - 39911.16
\item Loss secret3 - 39337.07
\item Loss Cover - 51310.23
\end{enumerate}

Currently, the above two results were taken for two and three secret images added to the cover image and then retrieved. As we increase the number of images the loss for all the values is expected to increase as more image features are being hidden in one single image. So, we need to find some threshold with respect to how many images can be added to the cover image to get decent results. We also have not explored the $\lambda_s$ value for the secret messages and the $\lambda_c$ for the cover image. This parameter may help incorrectly defining the loss equation and help in getting clearer results for the secret and encoded image. Currently for both the experiments we have taken the $\lambda_s$ and $\lambda_c$ as 1.

\section{Future Directions}
From the implementation perspective, we aim to,
\begin{enumerate}
\itemsep0em 
\item Increase the number of secret Images with lower loss.
\item Exploring $\lambda_s$ and $\lambda_c$ to see how it affects our results.
\item Use conditional decoders instead of multiple decoders.
\item We have used visual inspection as our primary evaluation metric, we can improve this by passing the encoded cover image through security software for pixel details verification.
\end{enumerate}

This project can enable exploration with steganography and, more generally, in placing supplementary information in images. Several previous methods have attempted to use neural networks to either augment or replace a small portion of an image-hiding system. We are trying to demonstrate a method to create a fully trainable system that provides visually excellent results in unobtrusively placing multiple full-size, color images into a carrier image. Extensions can be towards a complete steganographic system, hiding the existence of the message from statistical analyzers. This will likely necessitate a new objective in training and encoding smaller images within large cover images.

\begin{figure}[t]
\vskip 0.2in
\begin{center}
\centerline{\includegraphics[width=\columnwidth]{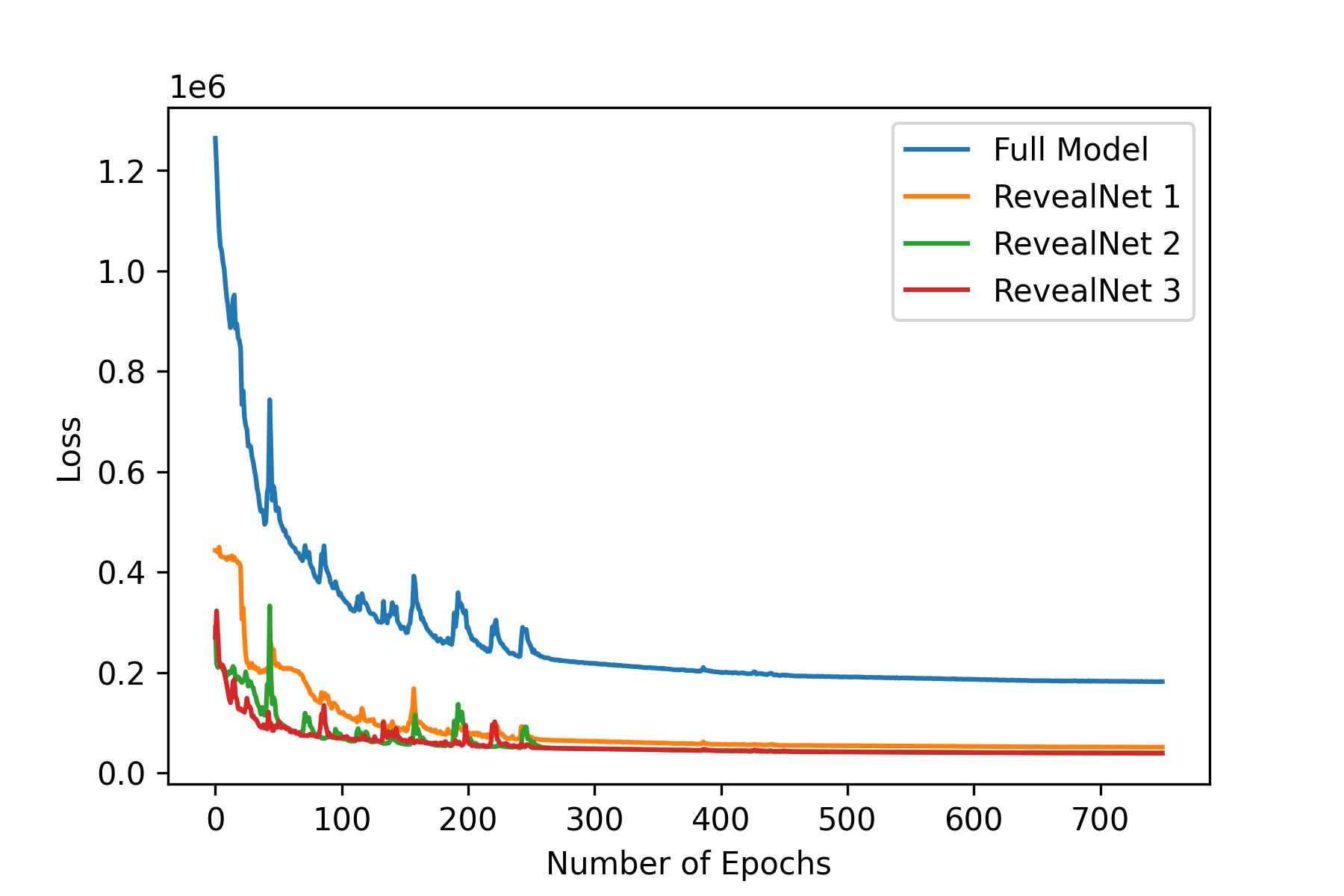}}
\caption{Graph of Overall Training Loss vs Number of Epochs till 750 epochs}
\label{fig:LossGraph}
\end{center}
\vskip -0.4in
\end{figure}

\section{Conclusion}
We got to understand Steganography in the image domain. The problem statement plays an important role in data security and we were able to gain more insights by reading various papers. Our implementation extended the single image steganography model proposed by \cite{NIPS2017_6802} by implementing multiple reveal networks corresponding to each secret image, as suggested by \cite{kreuk2019hide}. We were able to encode and decode up to three different secret images over a single cover image of similar size while maintaining decent min loss for secret images. Our cover image loss was higher though.

We relied heavily on visual perception for overall loss and didn't experiment with various types of losses which could have better suited for our model. A new method to validate the strength of the approach would help improve the results in the right direction. We plan to further extend the project with more secret images and work on various loss formulas. We plan to tweak the model to recover the encoded image with minimum loss. Our contribution of a deep neural network model in the field of Multi-Image Steganography can be extended further with GANs or even deeper neural networks.

\section{Acknowledgements}
We would like to express our heartfelt gratitude to Professor Bhiksha for providing us the topic and guiding us through the project. We would always like to thank our mentors Rohit Prakash Barnwal and Zhefan Xu for continuous support.

\FloatBarrier

\bibliography{example_paper}
\bibliographystyle{icml2020}

\end{document}